# ECAT: Event Capture Annotation Tool

**Tuan Do, Nikhil Krishnaswamy, James Pustejovsky**

Computer Science Department, Brandeis University

Waltham, Massachusetts USA

{tuandn,nkrishna,jamesp}@brandeis.edu

## Abstract

This paper introduces the Event Capture Annotation Tool (ECAT), a user-friendly, open-source interface tool for annotating events and their participants in video, capable of extracting the 3D positions and orientations of objects in video captured by Microsoft's Kinect® hardware. The modeling language VoxML (Pustejovsky and Krishnaswamy, 2016) underlies ECAT's object, program, and attribute representations, although ECAT uses its own spec for explicit labeling of motion instances. The demonstration will show the tool's workflow and the options available for capturing event-participant relations and browsing visual data. Mapping ECAT's output to VoxML will also be addressed.

## 1. Introduction

Much existing work in video annotation has focused on capturing objects from video in a purely two-dimensional format (i.e. tracking pixels) as in (Goldman et al., 2008), among others, or in capturing human body positioning in 3D for pose and gesture recognition (Kipp et al., 2014). We seek to wed these two types of capabilities by extracting the positions and orientations of objects and human body-rigs in video captured by the Microsoft Kinect®. These objects can be annotated as participants in a recorded motion event and this labeled data can then be used to build a corpus of *multimodal semantic simulations* of these events that can model object-object, object-agent, and agent-agent interactions through the durations of said events. This library of simulated motion events can serve as a novel resource of direct linkages from natural language to event visualization. We rely on the Kinect's capacity for body recognition and object tracking to produce output in the form of annotated object movement over time, allowing us to create an abstract representation of the denoted event.

The Kinect's depth field stream facilitates improved tracking of human movement, as reflected in the Kinect SDK's skeleton and face tracking performance (Livingston et al., 2012). The depth field provides a way to apply two-dimensional object tracking methods to a three-dimensional environment, which allows us to annotate captured video with a labeled event and its participants *with* their 3D positions throughout the event's duration. We can directly map from ECAT's output into VoxML, which was created specifically for modeling visualizations of objects and events. This mapping allows us to recreate the captured event instance in a simulated environment, and to begin compiling a library of labeled events and their participant objects simulated in 3D space, allowing in turn for the possibility of learning automatic discrimination of events from the motions of their participants.

ECAT is released as open source and it is available at https://github.com/tuandnvn/ecat.

## 2. Functionality

We use Kinect Sensor v2 for Windows which supports resolutions of up to HD 1920px × 1080px (RGB video) and 512px × 424px × 8 meters (depth). The latest SDK also supports 25 joint points of body tracking, and face tracking.

### 2.1. Capture and Input

For ECAT, we created our own capture and compression functionality rather than use the Kinect SDK's default functionality due to the large size of the resultant raw data files. Kinect capture automatically recognizes human bodies. Other objects may be manually marked by annotators. Once a video is captured and loaded, annotators may play it back and edit it. This may include removing an incorrectly recognized human body rig from the scene or cropping the video clip. The video clip may include frames beyond the interval of the captured event.

The default RGB color image and depth stream data are saved as separate video files. Body-tracking data is saved along with a scheme file specifying the name and index of every recognized joint in the body rig, how they are connected[1] and how they can be projected onto the RGB video. Additionally, users can import a property scheme file specifying what properties each object type can support, allowing them to modify the set of annotatable fields.

### 2.2. User Interface

Figure 1 shows the ECAT GUI. The various components are enumerated below.

1. Project management panel. Each project can hold multiple captured sessions.

2. Video display. For displaying either the color video or grayscale depth field video, and locating objects of interest in the scene—e.g., the table outlined in green in Figure 1.

3. Object annotation controller. Yellow time scrub bars show when each tracked object appears in the video.

---

[1] A human body rig is always a directed rooted tree whose nodes and edges form roughly the shape of a human stick figure.

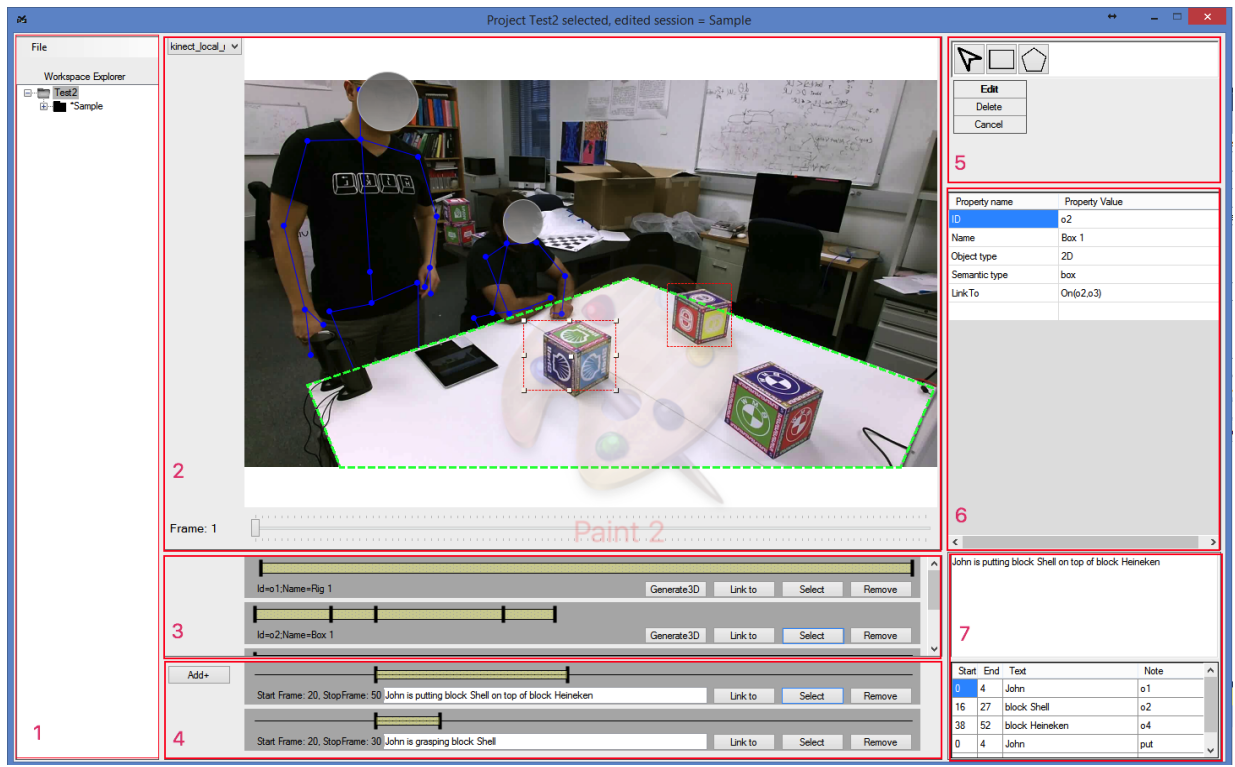

Figure 1: ECAT GUI. The left panel allows annotators to manage their captured and annotated sessions. Recognized human rigs display as blue skeletons. Marked object bounds display in color.

Black ticks mark frames where an annotator has drawn a bounding polygon around the object using the object toolbox (item 5). *Link to* links the selected object to another using a specified spatial configuration. *Generate3D* generates the selected object's tracking data using the depth field.

4. Event annotation controller. Time scrub bars here show the duration of a marked event. Users provide a text description for the event, or use *Link to* to link the selected event to another captured event as a subevent. ECAT supports marking events that comprise multiple non-contiguous segments. Due to space constraints not all annotated subevents are visible in this screenshot.

5. Object toolbox. Annotators can manually mark an in-video object with a bounding rectangle or arbitrary polygon. Marked bounds can be moved across frames as the object moves.

6. Object property panel. Data about a selected object shows here, such as ID and name.

7. Event property panel. The selected event's properties, including type and participants, show here, and the event can be linked to a VoxML event type.

Users can easily specify objects of interest in the scene, generate 3D tracking data, add or change object properties, and link them to VoxML objects. Events can be annotated with both natural language and a parametrized semantic markup, and linked to VoxML semantic programs.

## 2.3. Object Annotation

ECAT supports two ways of marking objects in a video. One is to import objects that have been automatically tracked using other libraries, such as human body rigs recognized by Kinect SDK. The other is to annotate locations of objects on the RGB video stream. Annotators mark the locations of objects at the beginning and end of an interval, and ECAT provides semi-automatic tracking using the depth field data and the iterative closest point method (Besl and McKay, 1992) to track the object's three-dimensional location. The output of the tracking algorithm can be either a point cloud or a parametric format if the object's shape can be approximated as a simple geometry (e.g., an orange or apple could be modeled as a spheroid, the tracking output being just the position of the object's center, and a radius).

An object's `objectType` field can be set to either *2D* or *3D*. Objects must be given an `ID`, `Name` and `semanticType`. We address usage of `semanticType` in Section 3..

Annotators may also mark relations between objects. For example, in Fig. 1, two blocks are on top of the table. Users can link a block object and the table object and specify the relation between the objects as "on," resulting in a predicate $On(Block\_1, Table)$ that is interpretable as a VoxML RE-LATION entity. Annotators could modify the available set or specify a different set of available relation predicates by importing a predicate scheme file. By default, ECAT supports the following binary predicates: $On$, $In$, $Attach\_to$, $Part\_of$.

## 2.4. Event Annotation

In principle, there are at least two ways to annotate an event associated with a video or video subinterval: (a) IDing an event type from an existing ontology or semantic resource, such as FrameNet (Baker et al., 1998); or (b) describing the event in natural language. We currently use the latter approach for filling an event's `text` field, but we are working toward incorporating ontologies with the `event` tag information, addressed in section 4..

As mentioned in section 2.2., ECAT allows annotation of event-subevent relations. Thus an overarching event may be annotated as *put*, but it contains the subevents *grasp*, *hold*, *move*, and *ungrasp*, which may overlap with some subsection of the main event and each other.

## 3. Links to VoxML

Entities modeled in VoxML can be objects, programs, attributes, relations, or functions. The VoxML OBJECT is used for modeling nouns, while PROGRAM is used for modeling events. The `semanticType` field of an object captured in ECAT, filled in with free NL input, can be linked to objects annotated in VoxML if objects with the specified label exist in the VoxML-based lexicon (the *voxicon*). An object of `semanticType`=*block* can be linked in a 3D scene to a VoxML object denoted by the lexeme *block*, linking the captured object to all the ontological and semantic data provided by the VoxML markup (e.g. an object marked with `semanticType`=*stack* will be assigned, in the ECAT-to-VoxML mapping, all the VoxML knowledge of what a "stack" is). Objects whose `objectType`=*3D* can then be placed or moved within such a scene according to the `Location` and `Rotation` tags from the video annotation. Thus ECAT annotation can be used to recreate an equivalent scene in a VoxML-based 3D environment.

The `semanticType` field of an annotated event can be attached to the motion of the objects in the scene that correspond to the event's participants. Thus, using the scene above as an example, the interaction of the *body_rig* object and the *block* objects can explicitly be marked as a *put* event, and the same object/agent motions can be recreated in a 3D scene, allowing for the creation of a linked dataset of annotated videos and procedurally generated scenes. This dataset could then be used to train machine-learning algorithms to discriminate motion events based on the motions of an event's respective participants in 3D space.

## 4. Output

Body rigs are saved as `objects` with `semanticType`=*body_rig*. They are ID'ed (`id`=*o1* as seen in Fig. 2) and can be given an alias for the user's ease (here *John*).

Annotated objects are treated similarly, assigned an ID, a name, and a semantic type. Here *o2* is the Shell logo block from Fig. 1. Object locations and relative spatial relations can be annotated by frame. At frame 1, *o2* is on the table (*o3*) while by frame 50, it has been put on the other block (*o4*), so the corresponding `LinkTo` tags are $On(o2, o3)$ and $On(o2, o4)$, respectively. By default, ECAT supports the relations $On$, $In$, $Part\_of$, and $Attach\_To$, where an object is in a parent-child relationship with another object, such as when a body rig's hand is carrying a block.

`annotations` denote events, with participants as referents (`refs`). In Fig. 3, *o1*, *o2*, and *o4* are `refs`, while *a1*'s `event`'s `semanticType`=*put*, marking the three above objects as the "put" event's participants. An `annotation`'s `superEvent` indicates super/subevent relationships, so that *a2*, a "grasp" event, is notated as a subevent of "put" *a1*.

Both `objects` and `annotations` can be mapped to VoxML representations, for instance as in Fig. 4 below, which shows a VoxML representation of *put*, an event annotated in Fig. 3.

Figure 2: Object output format.

Figure 3: Event annotation output format. Some subevent specifics are elided here for space.

Figure 4: VoxML for Fig. 3's *put* instance. *o1*, *o2*, and *o4* each point to that object's VoxML representation. $E_1$, $E_2$, and $E_3$ are mapped from annotated subevents, such as *grasp* in Fig. 3.

## 5. Conclusions

Event and action detection and recognition in video is receiving increasing attention in the scientific community, due to its relevance to a wide variety of applications (Ballan et al., 2011) and there have been calls for annotation infrastructure that includes video (Ide, 2013). We have presented here a tool that provides a user-friendly interface for video annotation that is able to capture a level of detail not provided by most existing video annotation tools, provides links to existing linguistic infrastructures, and is well suited for building a corpus of event-annotated multimodal simulations for use in the study of spatial and motion semantics (Pustejovsky and Moszkowicz, 2011; Pustejovsky, 2013).

For future annotation capabilities, we are planning on introducing links to existing semantic lexical resources, such as FrameNet, as well as event ontologies. More significantly, we are extending the ECAT environment to allow for annotation of much longer videos, encompassing multiple event sequences comprising narratives, including simultaneous or overlapping events that do not hold super/subevent relations between them but together make up a larger story (e.g. a man cooking dinner while a woman sets the table). This will entail enriching our specification to enable the markup of discourse connectives, linking the events in the narrative.

## Acknowledgements

This work is supported by DARPA Contract W911NF-15-C-0238 with the US Defense Advanced Research Projects Agency (DARPA) and the Army Research Office (ARO). Approved for Public Release, Distribution Unlimited. The views expressed are those of the authors and do not reflect the official policy or position of the Department of Defense or the U.S. Government. All errors and mistakes are, of course, the responsibilities of the authors.

## 6. Bibliographical References

Baker, C. F., Fillmore, C. J., and Lowe, J. B. (1998). The berkeley framenet project. In *Proceedings of the 17th international conference on Computational linguistics-Volume 1*, pages 86–90. Association for Computational Linguistics.

Ballan, L., Bertini, M., Del Bimbo, A., Seidenari, L., and Serra, G. (2011). Event detection and recognition for semantic annotation of video. *Multimedia Tools and Applications*, 51(1):279–302.

Besl, P. J. and McKay, N. D. (1992). Method for registration of 3-d shapes. In *Robotics-DL tentative*, pages 586–606. International Society for Optics and Photonics.

Goldman, D. B., Gonterman, C., Curless, B., Salesin, D., and Seitz, S. M. (2008). Video object annotation, navigation, and composition. In *Proceedings of the 21st annual ACM symposium on User interface software and technology*, pages 3–12. ACM.

Ide, N. (2013). An open linguistic infrastructure for annotated corpora. In *The People¿½s Web Meets NLP*, pages 265–285. Springer.

Kipp, M., von Hollen, L. F., Hrstka, M. C., and Zamponi, F. (2014). Single-person and multi-party 3d visualizations for nonverbal communication analysis. In *Proceedings of the Ninth International Conference on Language Resources and Evaluation (LREC), ELDA, Paris*.

Livingston, M. A., Sebastian, J., Ai, Z., and Decker, J. W. (2012). Performance measurements for the microsoft kinect skeleton. In *Virtual Reality Short Papers and Posters (VRW), 2012 IEEE*, pages 119–120. IEEE.

Pustejovsky, J. and Krishnaswamy, N. (2016). Voxml: A visual object modeling language. *Proceedings of LREC*.

Pustejovsky, J. and Moszkowicz, J. (2011). The qualitative spatial dynamics of motion. *The Journal of Spatial Cognition and Computation*.

Pustejovsky, J. (2013). Dynamic event structure and habitat theory. In *Proceedings of the 6th International Conference on Generative Approaches to the Lexicon (GL2013)*, pages 1–10. ACL.